\documentclass[%
reprint,
frontmatterverbose, 
amsmath,amssymb,
aps,
showkeys,
prstab,
floatfix,
]{revtex4-1}

\usepackage{graphicx}% Include figure files
\usepackage{dcolumn}% Align table columns on decimal point
\usepackage{bm}% bold math
\usepackage[draft]{todonotes}   % notes showed
\newcommand{\comment}[1]{}

\usepackage{hyperref}
\hypersetup{
    colorlinks=True,
    linkcolor=blue,
    urlcolor=blue,
    citecolor=gray,
}

\newcommand{\beginsupplement}{%
        \setcounter{table}{0}
        \renewcommand{\thetable}{S\arabic{table}}%
        \setcounter{figure}{0}
        \renewcommand{\thefigure}{S\arabic{figure}}%
     }

\begin{document}

\title{ \Large{Markerless tracking of user-defined features with deep learning}}

\author{{\bf Alexander Mathis$^{1,2,\dagger}$, Pranav Mamidanna$^{1}$, Taiga Abe$^{1,3}$,  Kevin M. Cury$^{3}$,\\ Venkatesh N. Murthy$^{2}$, Mackenzie Weygandt Mathis$^{1,4,\dagger^*}$ \& Matthias Bethge$^{1,5,6^*}$}}
\affiliation{1: Institute for Theoretical Physics, Werner Reichardt Center for Integrative Neuroscience, Eberhard Karls Universit\"at T\"ubingen, T\"ubingen, Germany}
\affiliation{2: Center for Brain Science \& Department of Molecular \& Cellular Biology \\ Harvard University, Cambridge, MA USA} 
\affiliation{3: The Mortimer B. Zuckerman Mind Brain Behavior Institute, Department of Neuroscience, Columbia University, New York, NY,  USA} 
\affiliation{4: The Rowland Institute at Harvard, Harvard University, Cambridge, MA USA} 
\affiliation{5: Max Planck Institute for Biological Cybernetics \\
\& Bernstein Center for Computational Neuroscience, T\"ubingen, Germany}
\affiliation{6: Center for Neuroscience and Artificial Intelligence, Baylor College of Medicine, Houston, TX USA} 
\affiliation{*co-senior authors}
\affiliation{$\dagger$corresponding authors: alexander.mathis@bethgelab.org \& mackenzie@post.harvard.edu}
\date{\today}

\begin{abstract}
{\bf Quantifying behavior is crucial for many applications in neuroscience. Videography provides easy methods for the observation and recording of animal 
behavior in diverse settings, yet extracting particular aspects of a behavior for further analysis can be highly time consuming. In motor control studies, humans or other 
animals are often marked with reflective markers to assist with computer-based tracking, yet markers are intrusive (especially for smaller animals), and the number and location of the 
markers must be determined \textit{a priori}. Here, we present a highly efficient method for markerless tracking based on transfer learning with deep neural networks that achieves excellent results with 
minimal training data. We demonstrate the versatility of this framework by tracking various body parts in a broad collection of experimental settings: mice odor trail-tracking, egg-laying behavior in drosophila, and mouse hand articulation in
a skilled forelimb task. For example, during the skilled reaching behavior, individual joints can be automatically tracked (and a confidence score is reported). Remarkably, even when a small number of frames are labeled ($\approx 200$), the algorithm achieves excellent tracking performance on test frames that is comparable to human accuracy.}
\end{abstract}

\maketitle

\begin{figure*}
\begin{center}
\includegraphics[width=\textwidth]{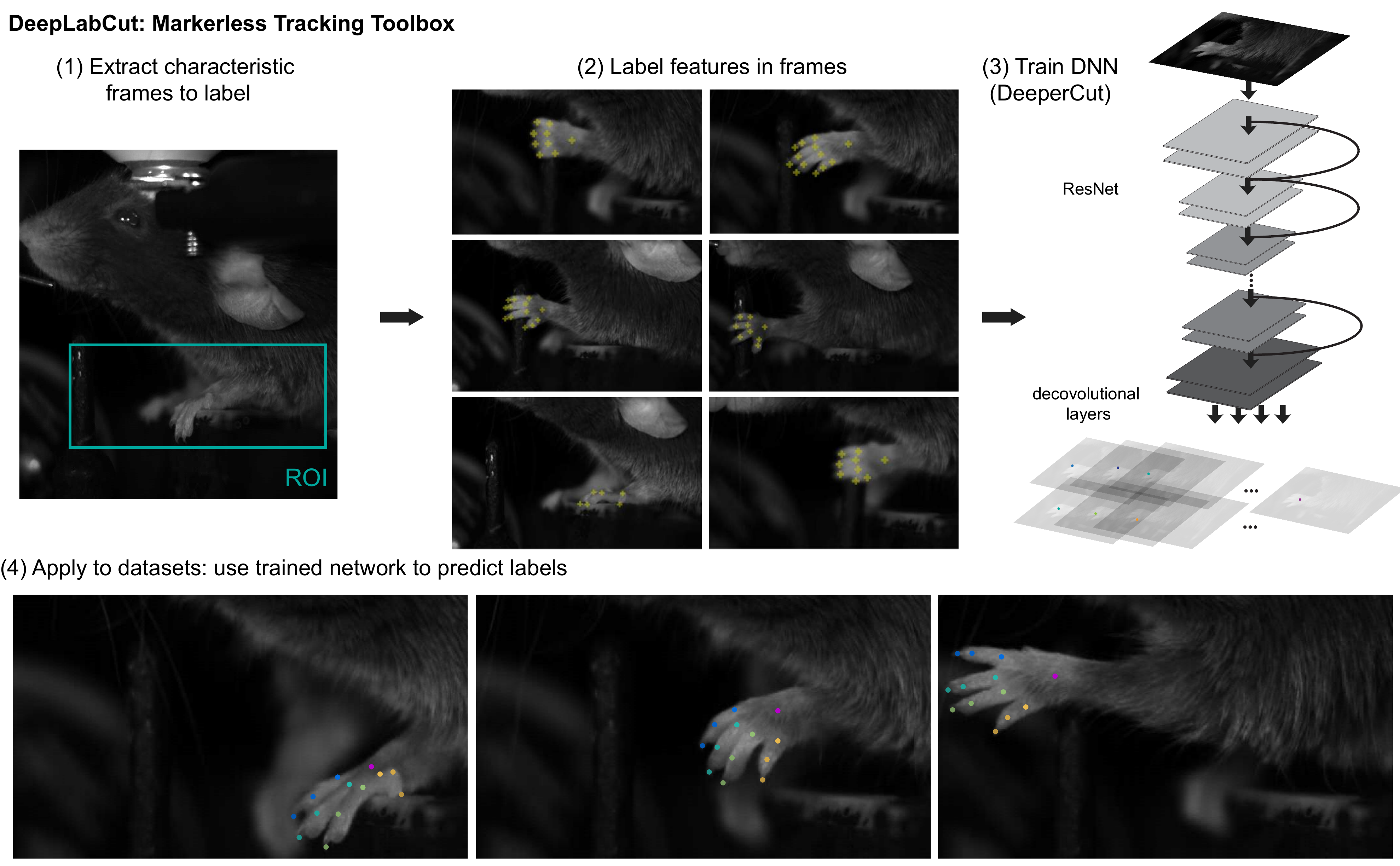}
\end{center}
\caption{{\bf Procedure for using the DeepLabCut Toolbox.} {\bf (1)} Training: extract images with distinct postures characteristic of the animal behavior in question. For computational efficiency, the region of interest (ROI) should be picked to be as small as possible while containing the behavior in question, which is reaching in the example. {\bf (2)} Manually localize (label) various body parts. Here, various digit joints and the wrist were selected as features of interest. {\bf (3)} Train a deep neural network (DNN) architecture to predict the body-part locations based on the corresponding image.  A distinct read-out layer per body part is generated to predict the probability that a body part is in a particular pixel. Training adjusts both readout and DNN weights. After training the weights are stored. {\bf (4)} The trained network can be used to extract the location of the body parts from videos. The images show the most likely body part locations for $13$ labeled body parts on the hand of a mouse.}
\label{fig:toolbox}
\end{figure*}

\section{Introduction}

Accurate quantification of behavior is essential for understanding the brain~\cite{Tinbergen1963,Bernstein1967,Krakauer2017}. Both within and beyond the field of neuroscience, there is a fruitful tradition of using cutting edge technology to study movement. Often, the application of new technology has the potential to reveal unforeseen features of the phenomena being studied, as in the case of Muybridge's famous photography studies in the mid 19th century, or modern high-speed videography that has revealed previously unknown motor sequences, such as ``tap dancing" in the songbird~\cite{Bernstein1967, Ota2015, Wade2016}. Historically, collected data was analyzed manually, which is a time-consuming, labor-intensive, and an error-prone process that is prohibitively inefficient at today's high rates of acquisition. Conversely, advances in computer vision have persistently inspired methods of data analysis to reduce human labor~\cite{Dell2014review,gomez2014big,Anderson2014}.

We are particularly interested in extracting the pose of animals, i.e. the geometrical configuration of multiple body parts. The gold standard for pose estimation in the field of motor control is the combination of video recordings with easily recognizable reflective markers applied to locations of interest, which greatly simplifies subsequent analysis and allows for tracking of body parts with high accuracy~\cite{Winter2009, Vargas-Irwin2010, Wenger2014, Maghsoudi2017}. However, such systems can be expensive, potentially distracting to
animals~\cite{Perez-Escudero2014, Nakamura2016}, and markers need to be placed before recording, which pre-defines the features that can be tracked. This mitigates one of the benefits of video data, i.e. it's low level of invasiveness. One alternative to physical markers is to fit skeleton/active contour models~\cite{deChaumont2012,Matsumoto2013, Perez-Escudero2014, Nakamura2016, Uhlmann2017}. These methods can work quite well and are fast, but require sophisticated skeleton models, which are difficult to develop and to fit to data, limiting the flexibility of such methods~\cite{Felzenszwalb2005,Toshev2014}. Another alternative is training regressors based on various computationally derived features to track particular body parts in a supervised way~\cite{Dollar2010, Perez-Escudero2014, Dell2014review, Machado2015Carey, Guo2015Hantman}. Training predictors based on features from deep neural networks also falls in this category~\cite{Krizhevsky2012, He_2016_CVPR}. Indeed, the best algorithms for challenging benchmarks in pose estimation of humans from images use deep features~\cite{Toshev2014, Wei16, pishchulin16cvpr, insafutdinov2016deepercut,  Feichtenhofer2017,insafutdinov2017cvpr}. This suggests that deep learning architectures should also greatly improve the accuracy of pose estimation for lab applications. However, the labeled data sets for these benchmarks are rather large (e.g. $\approx 25,000$ in the MPII Human Pose data set~\cite{andriluka20142d}), which may render deep learning approaches infeasible as efficient tools at the scale of interest to neuroscience labs. Yet, due to transfer learning~\cite{donahue2014decaf,Yosinski2014,Goodfellow2016,kummerer2016deepgaze} we will show that this need not be the case. 

Here, we demonstrate that by capitalizing on state-of-the-art methods for detecting human limb configurations, we can achieve excellent performance on pose estimation problems in the laboratory setting with minimal training data. Specifically, we investigated the feature detector architecture from DeeperCut~\cite{insafutdinov2016deepercut,pishchulin16cvpr}, one of the best performing pose estimation algorithms, and demonstrate that a small number of training images ($\approx 200$) can be sufficient to train this network to within human-level labeling accuracy. This is possible due to transfer learning: the feature detectors are based on extremely deep neural networks, which were pre-trained on ImageNet, a massive data set for object recognition~\cite{He_2016_CVPR}. By labeling only a few hundred frames one can train tailored, robust feature detectors that are capable of localizing  a variety of experimentally relevant body parts. We illustrate the power of this approach by tracking the snout, ears, and tail-base during an odor-guided navigation task, multiple body parts of a fruit fly behaving in a 3D chamber, as well as joints of individual digits during a reaching task.

\begin{figure*}
\makebox[\linewidth]{\includegraphics[width=\textwidth]{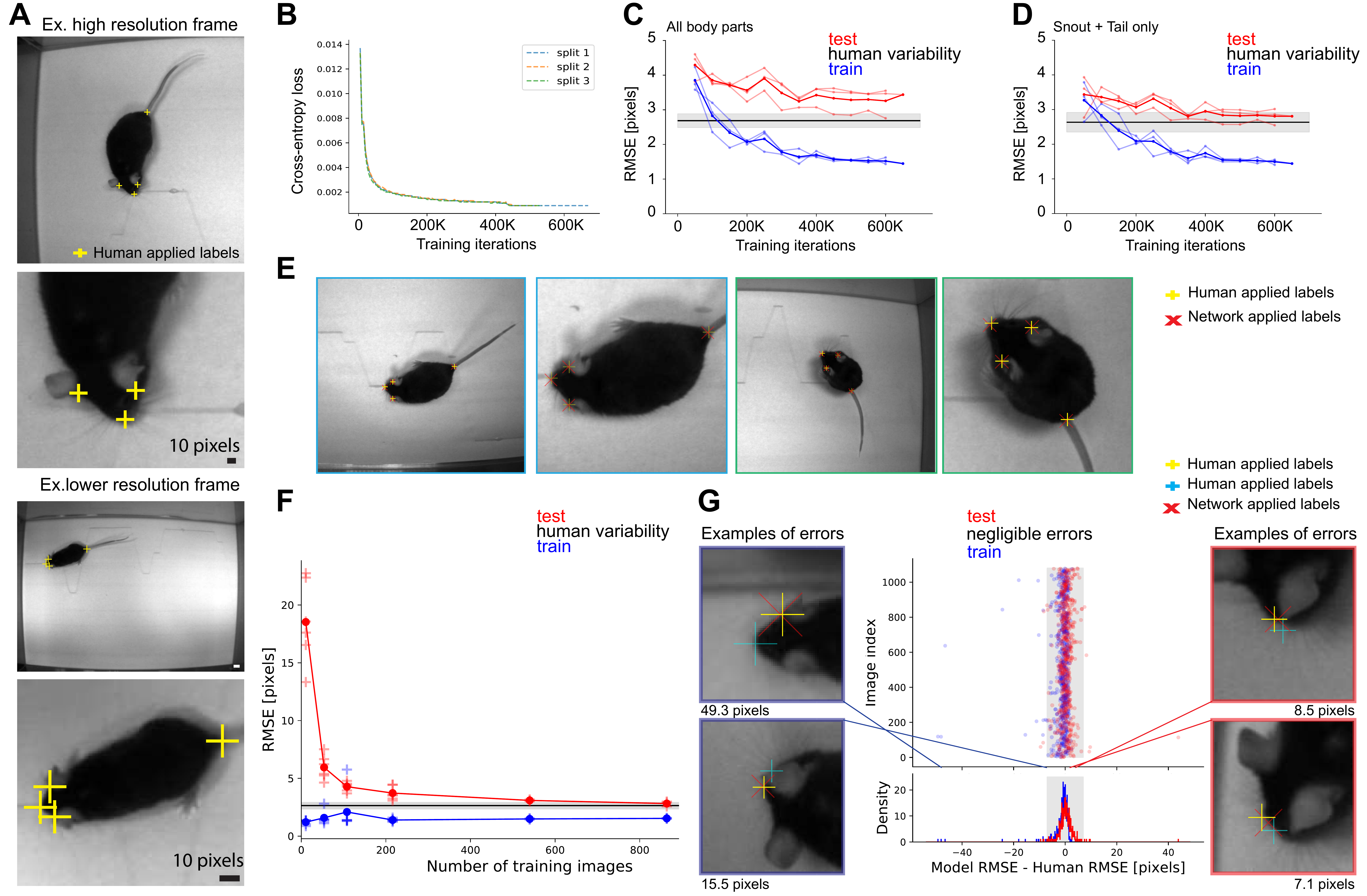}}
\caption{{\bf Evaluation during Trail-Tracking.} {\bf A}: Two example frames with close-ups, showing snout, ears and tail base labeled by a human. The odor-trail as well as reward drops are visible under infra-red light and introduce a time-varying visual background. {\bf B}: Cross-entropy loss for $3$ splits when training with $80\%$ of the $1\,080$ frames. {\bf C}: Corresponding root mean square error (RMSE) between the human and the predicted label on training and test images for those $3$ splits evaluated every $50\,000$ iterations ($80\%/20\%$ splits). The average of those individual curves is also shown (thick line). Human variability as well as $95\%$ confidence interval are depicted in black/gray.  {\bf D}: Corresponding RMSE when evaluated only for snout and tail base. Thus, the algorithm reaches human level variability on a test set comprising $20\%$ of all images. {\bf E}: Example body part prediction for two frames, which were not used during the training (of one split in C). Prediction closely matches human annotator's labels. {\bf F}: RMSE for snout and tail for several splits of training and test data compared to RMSE of a human scorer. Each split is denoted by a cross, the average by a dot. For $80\%$ of the data the algorithm achieves human level accuracy on the test set (D). As expected, test RMSE increases for fewer training images. Around $100$ frames are enough to provide average tracking performance ($\leq 5$ pixel accuracy). {\bf G}: Snout-RMSE comparison between human and model per image for one split with $50\%$ training set size. Most RMSE differences are small with few outliers. The two extreme errors (on the left) are due to labeling errors across trials by the human.}
\label{fig:TrackingMainFig}
\end{figure*}

\section{Results}

DeeperCut achieves outstanding performance on multi-human pose detection benchmarks~\cite{insafutdinov2016deepercut}. However, to achieve this performance its neural network architecture has been trained on thousands of labeled images. Here we focus on a subset of DeeperCut: its feature detectors, which are variations of Deep Residual Neural Networks (ResNet)~\cite{He_2016_CVPR} with readout layers that predict the location of a body part (see Methods, Figure~\ref{fig:toolbox}). To distinguish the feature detectors from the full DeeperCut, we refer to this autonomous portion as DeepLabCut. In this paper, we (1) evaluate the performance of DeepLabCut for posture tracking in various laboratory behaviors, (2) investigate the amount of required training data for good generalization, and (3) provide an open source toolbox based on DeeperCut that is broadly accessible to the neuroscience community (\url{https://github.com/AlexEMG/DeepLabCut}). 

\subsection{Benchmarking} 

Analyzing videos taken in a dynamically changing environment can be challenging. Therefore, to test the utility of our toolbox, we first focused on an odor-guided navigation task for mice. Briefly, mice run freely on an `endless' paper spool that includes an adapted ink-jet printer to deliver odor trails in real-time as a mouse runs and tracks trails (further details and results will be published elsewhere). The video captured during the behavior poses several key challenges: inhomogeneous illumination, transparent side walls that appear dark, shadows around the mouse from overhead lighting, distortions due to a wide-angle lens, the mouse often crosses the odor trail, and rewards are often directly in front of its snout ( which influences its appearance). Yet, accurately tracking the snout as a mouse samples the ``odorscape" is crucial for studying odor-guided navigation. 

\begin{figure*}[ht!]
\begin{center}
\includegraphics[width=\textwidth]{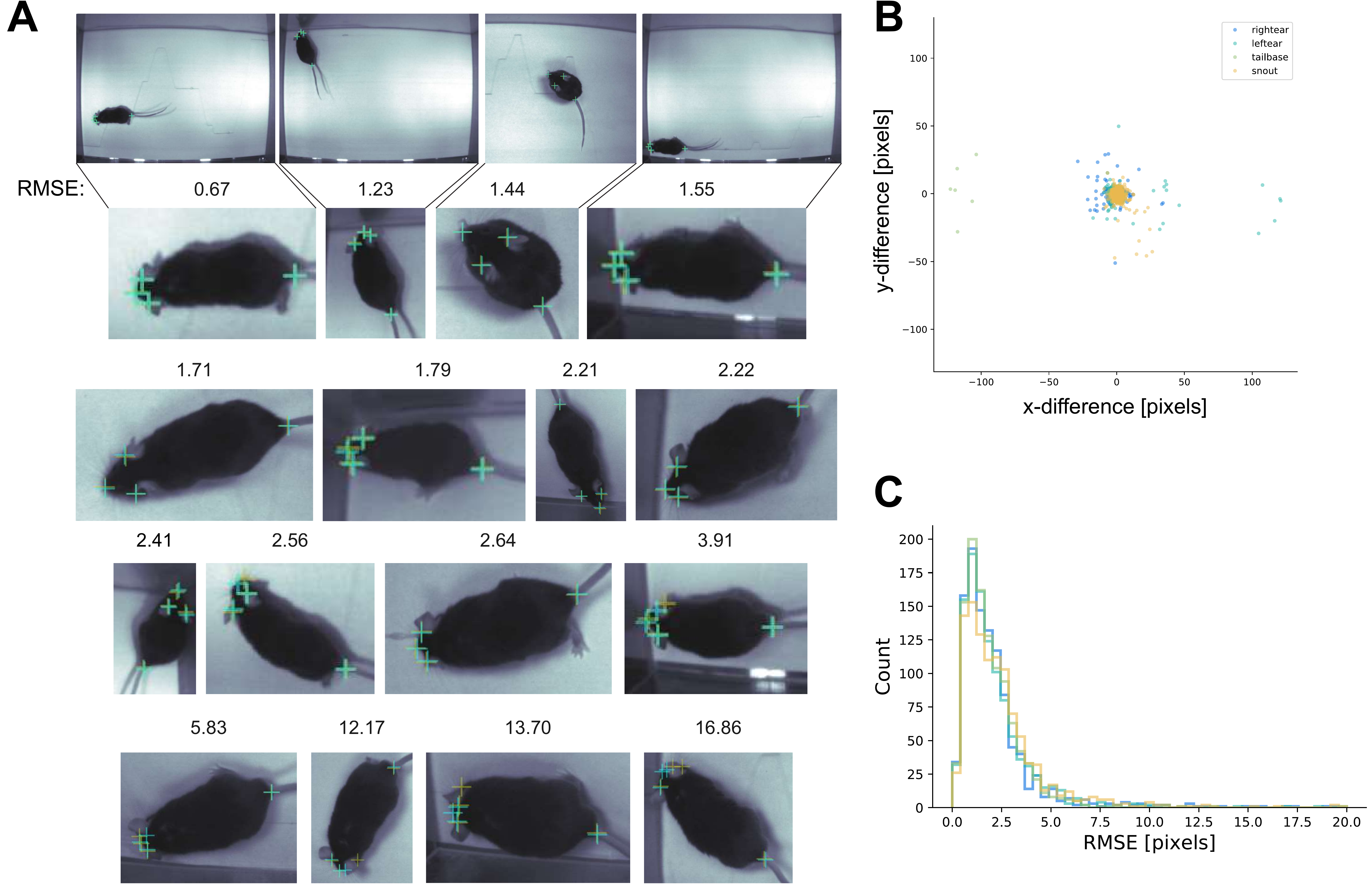}
\end{center}
\caption{{ \bf Illustration of labeled data set}. {\bf A:} Example images with human applied labels (first and second trial by the same labeler colored in yellow and cyan) illustrating the variability. Top row shows full frames that were labeled and illustrate typical examples of the $1\,080$ random frames, which comprise the data set. Rows below are cropped for visibility of the labels and indicate the average RMSE in pixels across all body parts above the image. The scorer was highly accurate, as illustrated by B. {\bf B:} x-axis and y-axis difference in pixels between the first - second trial. Only a few labels strongly deviate between the two trials. Most errors are smaller than $5$ pixels as can be seen in the histogram of trial-by-trial labeling errors (cropped at $20$ pixels) ({\bf C}).}
\label{fig:Dataset}
\end{figure*}

First, we extracted $1,080$ distinct frames from multiple videos (across two cameras and $7$ different mice, see Methods) and manually labeled the snout, left ear, right ear, and tail base in all frames (Figure \ref{fig:TrackingMainFig}A, Figure~\ref{fig:Dataset}). In order to facilitate comparisons to ground truth and to quantify the robustness of predictors, we estimated variability (root mean square error; RMSE) within one human labeler by comparing two distinct label sets of the same data. We found the average variability for all body parts to be very small $2.69 \pm 0.1$ pixels (mean $\pm$ s.e.m. $n=4,320$ body part image pairs; Figure~\ref{fig:Dataset}, see Methods), which is less than the width of the mouse's snout in low resolution camera frames (Figure~\ref{fig:TrackingMainFig}A). The RMSE across two trials of annotating the same images is referred to as `human variability' (note that the variability differs slightly across body parts). 

To quantify the feature detector's performance, we randomly split the data into a training and test set ($80\%/20\%$) and evaluated the performance of DeepLabCut on test images across all body parts (Figure~\ref{fig:TrackingMainFig}B, C) and in a sub-set of body parts (snout/tail)(Figure~\ref{fig:TrackingMainFig}D). Unless otherwise noted, we always trained (and tested) with the labels from the first set of human labels. The test RMSE for different training/test set splits achieved average human variability (Figure~\ref{fig:TrackingMainFig}D). Thus, we found that when trained with $80\%$ of the data the algorithm achieved human-level accuracy on the test set for detection of the snout and the tail base (Figure~\ref{fig:TrackingMainFig}D, E).

Next, we systematically varied the size of the training set and trained $30$ distinct networks ($3$ splits for $50\%$ and $80\%$ training set size; $6$ splits for $1$, $5$, $10$ and $20\%$ training set fraction). As expected, the test error increases for decreasing number of training images (Figure~\ref{fig:TrackingMainFig}F). Yet remarkably, the test RMSE only slowly attenuates from $80\%$ training set fraction to $10\%$, where one still achieves an average pixel error of less than $5$ pixels. Such average errors are on the order of the size of the snout in the low resolution camera (around $5$ pixels) and much smaller than the size of the snout in higher resolution camera (around $30$ pixels). Thus, we found that even $100$ frames are enough to achieve excellent generalization. 

\begin{figure}
\includegraphics[width=.98\columnwidth]{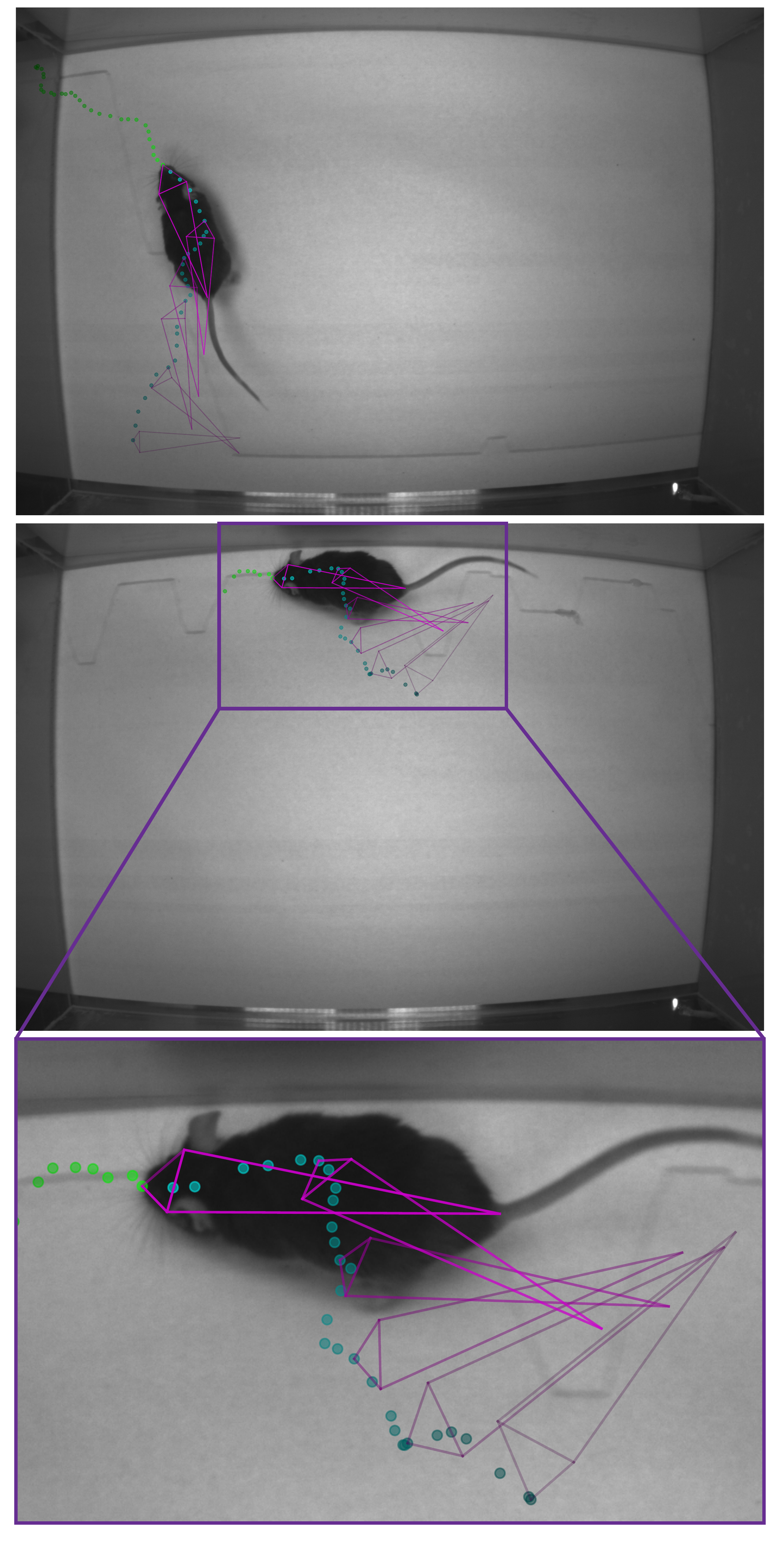}
\caption{{\bf A trail-tracking mouse.} Green and cyan dots show $30$ future and past (respectively) snout positions of the snout during trail-tracking, each $33.\overline{3}$ ms apart. The body postures of the the snout, ears and tail base at various past time points are depicted as magenta rhombi. Together those four points capture the body and head orientation of the mouse and illustrate the swinging head movements. The printed odor trail is visible in gray (see video at \url{http://www.mousemotorlab.org/deeplabcut}).}
\label{fig:trailtracking}
\end{figure}

Since the RMSE is computed as an average across images, we next checked if there were any systematic differences across images by comparing the human variability across the two splits vs. the model variability (trained with the first set of human labels; Figure~\ref{fig:TrackingMainFig}G: data for $1$ split with a $50\%$ training set size). We found that both the human and the algorithm produced only a few outliers, but no systematic error was detected (see Figure~\ref{fig:TrackingMainFig}G for examples). 

Thus far, we used a part detector based on the $50$ layer deep ResNet-$50$~\cite{insafutdinov2016deepercut,He_2016_CVPR}. We also trained deeper networks with $101$ layers and found that both the training and testing errors decreased slightly, suggesting that the performance can be further improved if required (Average test RMSE for $3$ identical splits of $50\%$ training set fraction: ResNet-$50$: $3.09 \pm 0.04$, ResNet-$101$: $2.90 \pm 0.09$ and ResNet-$101$ with intermediate supervision: $2.88 \pm 0.06$, pixel mean $\pm$ s.e.m.; see Figure~\ref{fig:Crossvalidation}A). 

Overall, given the robustness and the low error rate of DeepLabCut even with small training sets, we find this to be a useful tool for studies such as odor-guided navigation. For example, Figure~\ref{fig:trailtracking} recapitulates a salient signature of the tracking behavior, namely that rodents swing their snout across the trail~\cite{Khan2012}. Knowing the location of the ears as well as the tail is also important to computationally assess the orientation of the mouse (Figure~\ref{fig:trailtracking}). Furthermore, having an automated pose estimation algorithm as presented will be crucial for video-rich experiments (such as this one). 

\subsection{Generalization \& transfer learning}

\begin{figure}
{\includegraphics[width=\columnwidth]{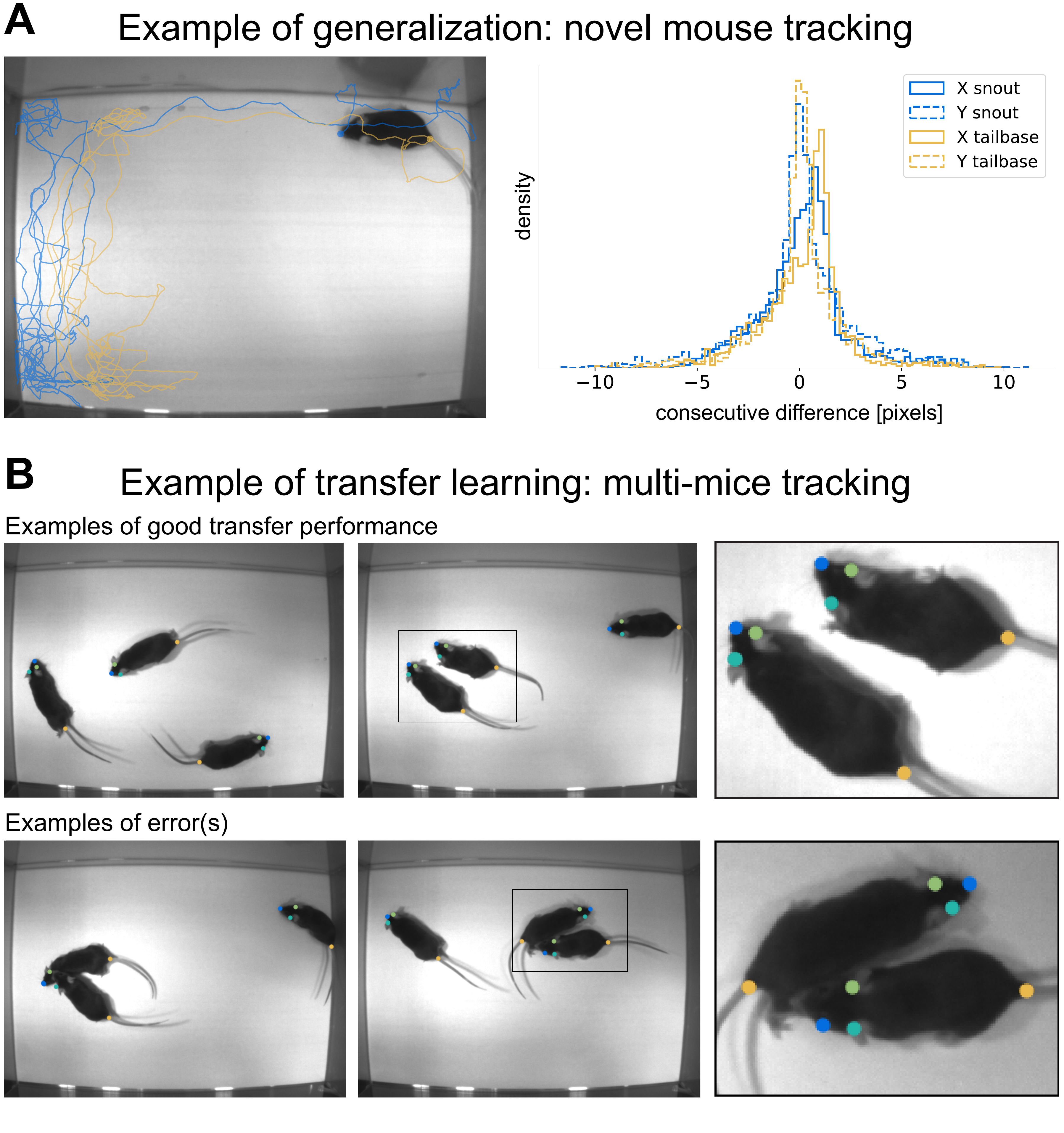}}
\caption{{\bf Generalization}: {\bf A}: Frame-by-frame extracted snout and tail base trajectory in a novel mouse (not part of training set) during trail tracking on moving paper ground. Continuity of the trajectory suggests accurate training, which is also confirmed by histogram of pairwise, frame-by-frame differences in x- and y-coordinates. {\bf B}: Body part predictions for images with multiple mice. Predictions of a network which was trained with images only containing a single mouse. It readily detects body parts in images with multiple (novel) mice (top), unless they are occluding each other (bottom). Additionally, these mice are younger than the ones in the training set (and have a different body shape).}
\label{fig:socialmicetracking}
\end{figure} 

We have demonstrated that DeepLabCut can accurately detect body parts across different mice, but how does it generalize to novel scenarios? Firstly, we found that DeepLabCut generalizes to novel mice during trail tracking (Figure~\ref{fig:socialmicetracking}A). Secondly, we tested if the trained network could identify multiple body parts across multiple mice within the same frame (transfer learning). Remarkably, although the network has only been trained with images containing a single mouse, it could detect all the body parts of each mouse in images with multiple interacting mice. Although not error-free, we found that the model performed remarkably well in a social task (three mice interacting in an unevenly illuminated open field, Figure~\ref{fig:socialmicetracking}B). The performance of the body part detectors could be improved by training the network with training images that include multiple mice with occlusions, and/or by training image-conditioned pairwise terms between body parts to harness the power of multi human pose estimation models~\cite{insafutdinov2016deepercut} (see discussion). Nonetheless, this example of multi-mice tracking illustrates that even the single-mouse trained feature detectors alone can readily transfer to extensions, as would be useful in studies of social behaviors~\cite{Dell2014review,Li2017,robie2017machine}.

\subsection{The power of end-to-end training}

\begin{figure}
{\includegraphics[width=.96\columnwidth]{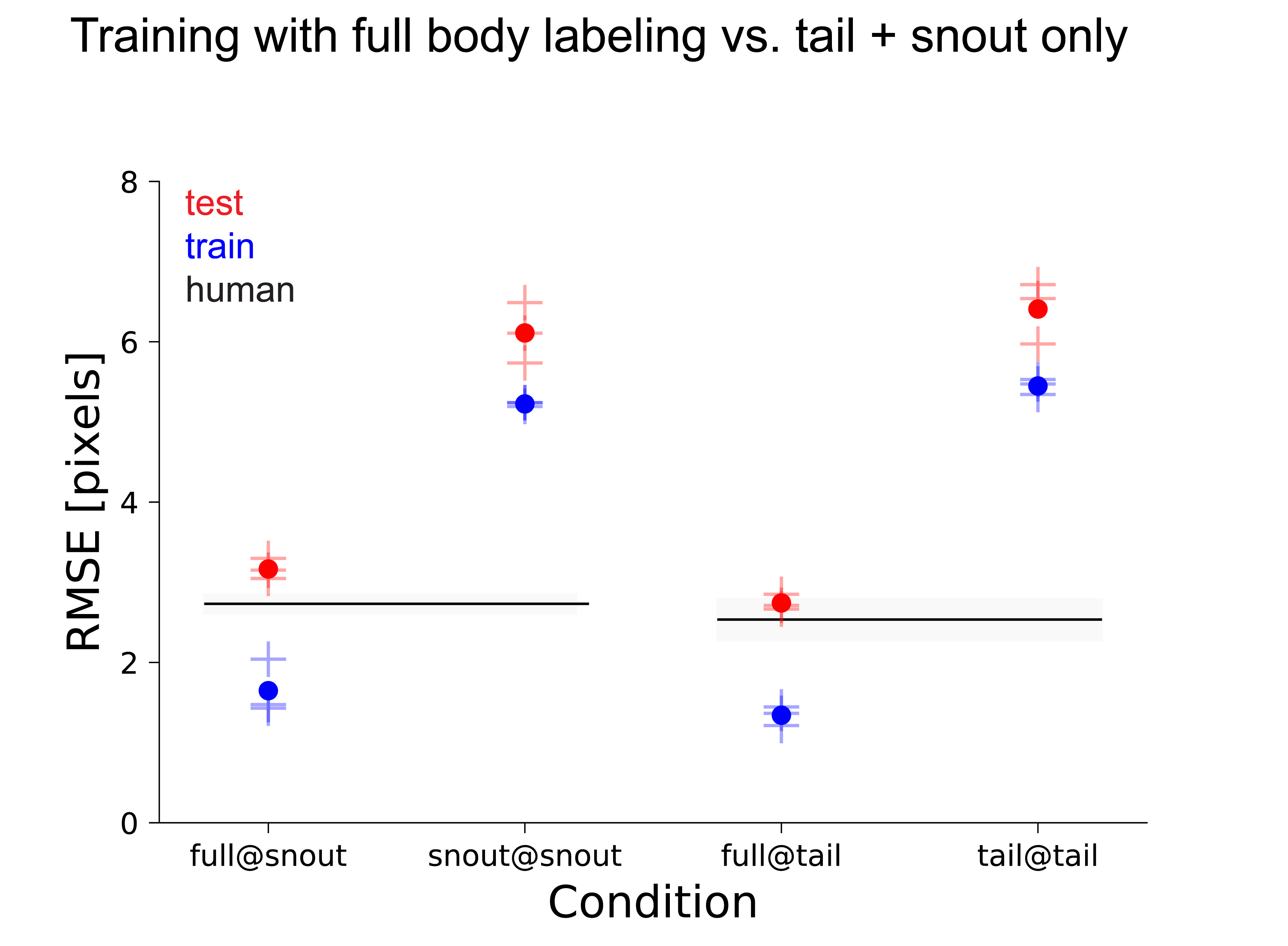}}
\caption{{\bf End-to-end training}: We trained `specialized' networks with only the snout or tail labels, respectively. We compare the RMSE against the full model that was trained on all body parts, but is also only evaluated on the snout or tail, respectively (i.e. ``full@snout" means the model trained with all body parts (full), and RMSE evaluated on the snout). Training (blue) and test (red) performance for the full model and specialized models trained with the same three splits $50\%$ of the data (crosses) and average RMSE (dots). Although all networks have exactly the same information about the location of the snout or tail during training, the network that also received information about the other body parts outperforms the `specialized' networks.                                            }
\label{fig:bodypartgeneralization}
\end{figure}

\begin{figure*}
\begin{center}
\includegraphics[width=\textwidth]{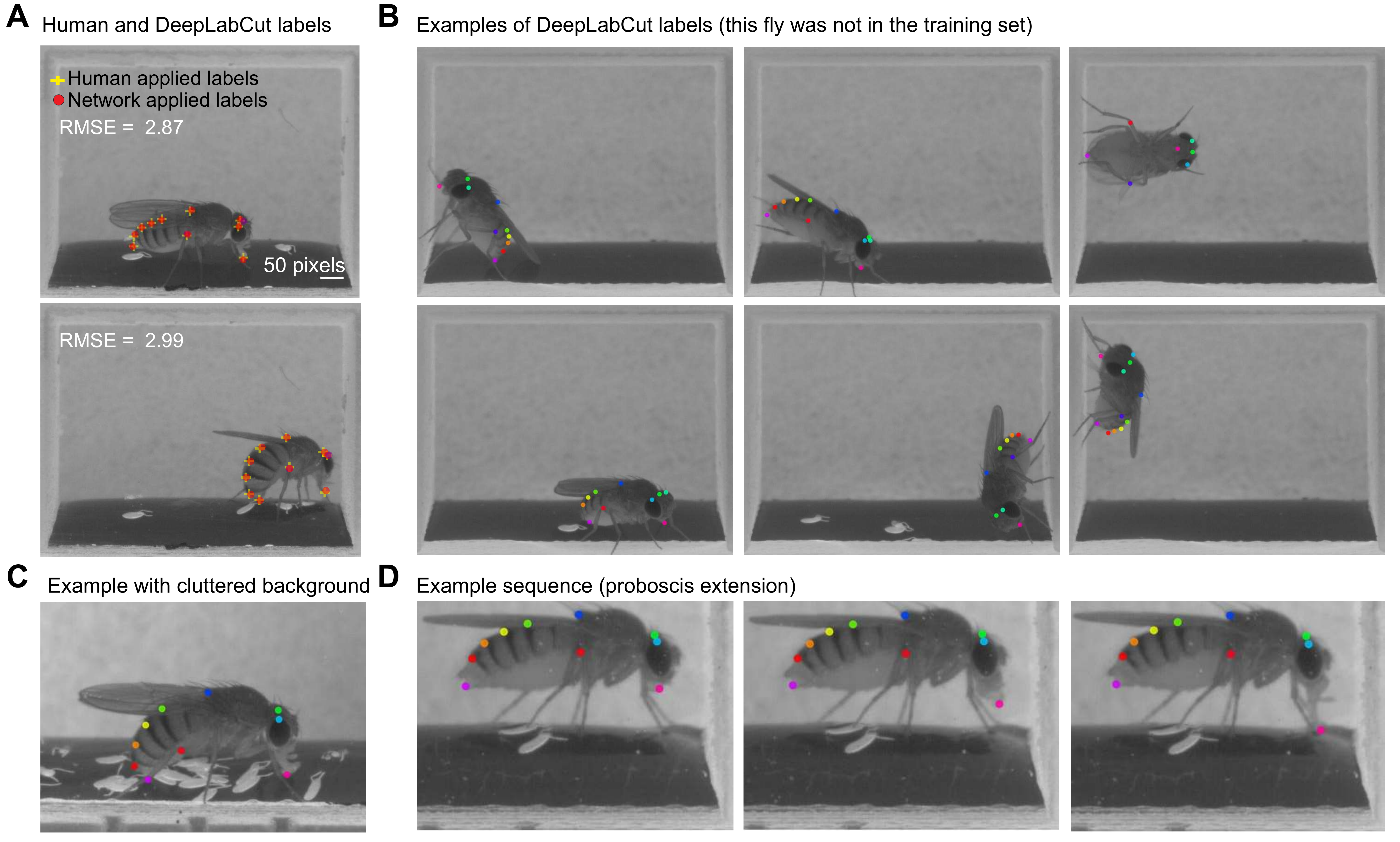}
\end{center}
\caption{{\bf Markerless Tracking of Drosophila.} {\bf A}: Example body part predictions closely match with human annotator’s labels, shown for two frames that were not used in the training set ($95\%$ training set size). {\bf B}: Example frames and body part predictions for a fly that was not part of the training data in various postures and orientations. {\bf C}: Example labels against a cluttered background comprised of numerous laid eggs. {\bf D}: Example sequence of proboscis extension being automatically and accurately tracked (same fly as B). See video at \url{http://www.mousemotorlab.org/deeplabcut}.}
\label{fig:Drosophila}
\end{figure*}

Due to the architecture of DeeperCut, the deconvolution layers are specific to each body part, but the deep pre-processing network (ResNet) is shared (Figure~\ref{fig:toolbox}). We hypothesized that this architecture can facilitate the localization of one body part based on other labeled body parts. To test this hypothesis, we examined the performance of networks trained with only the snout/tail data while using the identical $3$ splits of $50\%$ training data as in Figure~\ref{fig:TrackingMainFig}B. We found that the network that was trained with all body part labels simultaneously outperforms the specialized networks nearly twofold (Figure~\ref{fig:bodypartgeneralization}). This result also demonstrates that training the weights throughout the whole network in an end-to-end fashion rather than just the readout weights substantially improves the performance. This further highlights the advantage of such deep learning based models over approaches with fixed feature representations, which cannot be refined during training. 

\subsection{Drosophila in 3D behavioral chamber}

To further demonstrate the flexibility of the DeepLabCut toolbox, we tracked the bodies of freely behaving fruit flies (Drosophila) exploring a small cubical environment in which one surface contained an agar-based substrate for egg laying. Freely behaving flies readily exhibit many orientations, and also frequent the walls and ceiling. When viewed from a fixed perspective, these changes in orientation dramatically alter the appearance of flies as the spatial relationship of body features change, or as different body parts come into or out of view. Moreover, reliably tracking features across an entire egg-laying behavioral session could potentially be challenging to DeepLabCut owing to significant changes in the background (the accumulation of new eggs or changes in the agar substrate appearance due to evaporation). 

To build towards an understanding of the behavioral patterns that surround egg-laying in an efficient way, we chose $12$ distinct points on the body of the fly and labeled $589$ frames of diverse orientation and posture from six different animals, labeling only those features that were visible within a given frame. We labeled four points on the head (the dorsal tip of each compound eye, the ocellus, and the tip of the proboscis), the posterior tip of the scutellum on the thorax, the joint between the femur and tibia on each metathoracic leg, the abdominal stripes on the four most posterior abdominal segments (A$3$-A$6$), and the ovipositor.

\begin{figure*}
\begin{center}
\includegraphics[width=.98\textwidth]{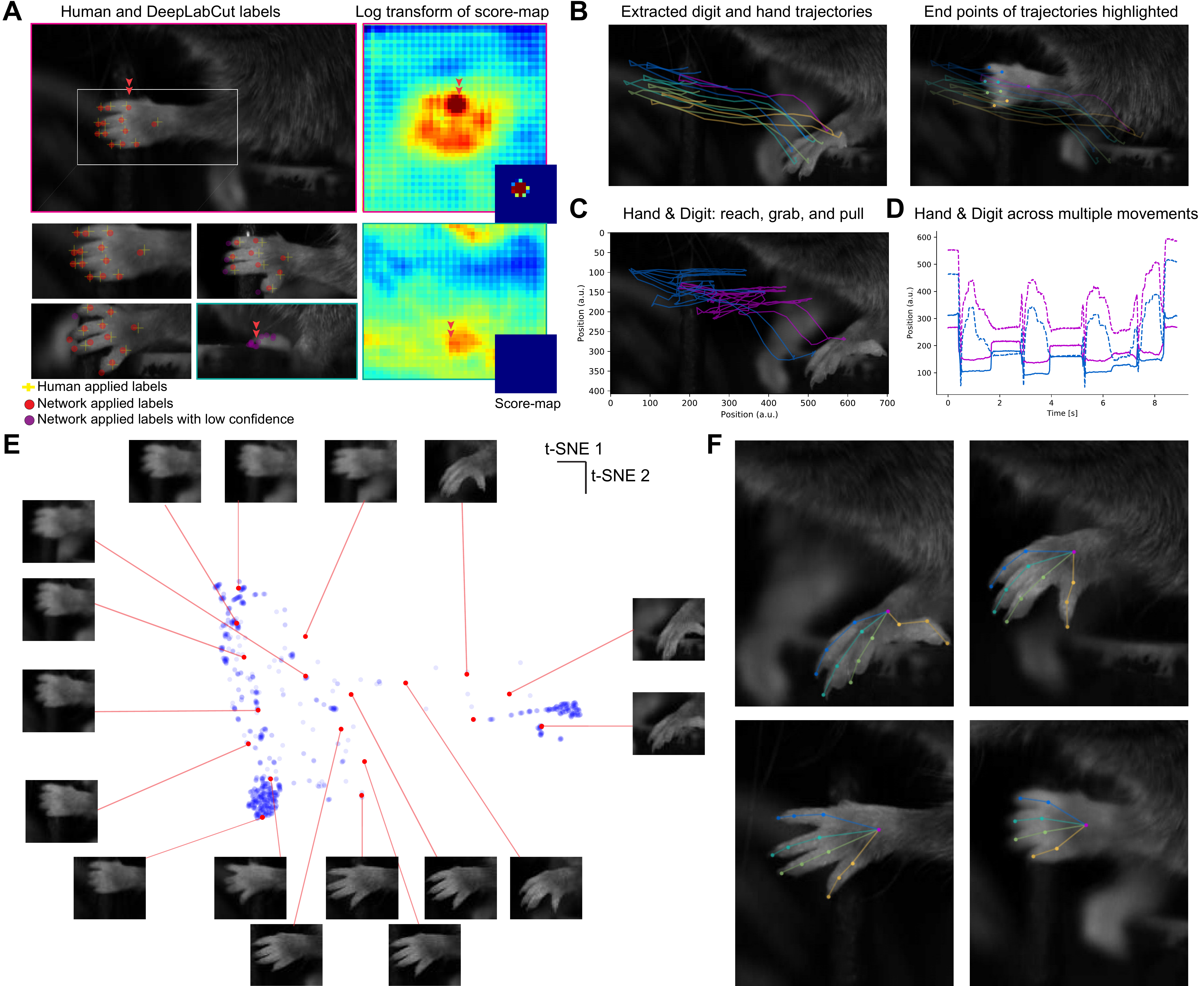}
\end{center}
\caption{{\bf Markerless Tracking of Digits.} {\bf A}: Example test images labeled by human (yellow) and automated labeling (red/purple): the most likely joint is labeled in red if the probability of the peak in the score-map (see below) is larger than 10\% at the peak, and purple otherwise. Here DeepLabCut was trained with only $141$ images. In images where targets were occluded, the model typically reduces its confidence in the label position (purple), including in frames where the human was not confident enough to apply labels (bottom right) or when the digit tips are not visible (upper right). Score-maps are shown for two example labels (highlighted by red arrows), and score-maps are to the right of the image. For better visibility log-transformed score-maps are also shown. {\bf B}: Example analysis of trajectories after automated labeling. {\bf C}: Extracted reach and pull trajectories from the wrist (purple), and digit 1 (blue) for $3$ pulls {\bf D}: Trajectories in {\bf C}. Dashed lines are the X-axis and solid lines represent the Y-axis. {\bf E}: Based on the predicted wrist location, we extracted images of the hand from the video of one behavioral session and performed dimensionality reduction by t-SNE embedding of those images. The blue point cloud shows the 2D-embedding with several images corresponding to the red highlighted coordinates. This figure illustrates the richness of hand postures during reaching and pulling of the mouse. {\bf F}: Labeled body parts with connecting edges giving rise to `skeleton' of hand (see video at \url{http://www.mousemotorlab.org/deeplabcut}).}
\label{fig:Reaching}
\end{figure*}

We trained DeepLabCut with $95\%$ of the data and found a test error of $4.17$ $\pm$ $0.32$ pixels (mean $\pm$ s.e.m.; corresponding to an average training error of $1.39$ $\pm$ $0.01$ pixels, n=$3$ splits, mean $\pm$ s.e.m.). Figure~\ref{fig:Drosophila}A depicts some example test frames with human and network applied labels. Generalization to flies not used in the training set was excellent and the feature detectors are robust to changes in orientation (Figure~\ref{fig:Drosophila}B) and background (Figure~\ref{fig:Drosophila}C). Although fly bodies are relatively rigid, which simplifies tracking, there are exceptions. For instance, the proboscis dramatically changes its visual appearance during feeding behaviors. Yet, the feature detectors can resolve fast motor movements such as the extension and retraction of the proboscis (Figure~\ref{fig:Drosophila}D). Thus, DeepLabCut allows accurate extraction of low-dimensional pose information from videos of freely behaving Drosophila.

\subsection{Digit Tracking During Reaching}

To further illustrate the versatility and capabilities of DeepLabCut we tracked segments of individual digits of a mouse hand (Figure~\ref{fig:toolbox} and \ref{fig:Reaching}A). We recently established a head-fixed, skilled reaching paradigm in mice~\cite{Mathis2017}, where mice grab a 2-degree of freedom joystick and pull it from a virtual start location to a target location. While the joystick allows for spatially accurate measurement of the joystick (hand) position during the pull with high temporal precision, it neither constrains the hand position on the joystick, nor provide position during the reaches or between pulls (when the mice might or might not hold the joystick). Placing markers is difficult as the mouse hand is a small, and a highly complex structure with multiple bones, joints, and muscles. Moreover, it is intrusive to the mice and can disrupt performance. Therefore, in principle, markerless tracking is a promising approach for analyzing reaching dynamics. However, tracking is challenging due to the complexity of possible hand articulations, as well as the presence of the other hand in the background. The images are also subject to very different image statistics, making this task well posed to highlight the generality of our DeepLabCut toolbox. 

We labeled $13$ points per frame: $3$ points per visible digit ($4$ digits): the digit tip, one mid digit joint, the base of the digit joint (these roughly correspond to the proximal interphalangeal joint (PIP) and the metacarpophalangeal joint (MCP), respectively), as well as $1$ point at the base of the hand (wrist). Remarkably, we found that by  using just $141$ training frames we achieved an average test error of $5.21 \pm 0.28$ pixels (mean $\pm$ s.e.m.; corresponding to average training error $1.16 \pm 0.03$ pixels, $n=3$ splits, mean $\pm$ s.e.m.). For reference, the width of a single digit is $\approx15$ pixels. Figure~\ref{fig:Reaching}A depicts some example test frames. We believe that this application of hand pose estimation highlights the excellent generalization performance of DeepLabCut despite training with only a few images. 

So far we have shown that the body part estimates derived from DeeperCut are highly accurate. But, in general, especially when sieving through massive data sets, the end user would like to have each point estimate accompanied by a confidence measure of the label location. The location predictions in DeeperCut are obtained by extracting the most likely region, based on a scalar field that represents the probability that a particular body part is in a particular region. In DeeperCut these probability maps are called ``score-maps", and predictions are generated by finding the point with the highest probability value (see Methods). The amplitude of the maximum can be used as a confidence read-out to examine the strength of evidence for individual localizations of the individual parts to be detected. For instance, the peak probability of the digit tip is low when the mouse holds the joystick (which causes occlusions of the digit tips). Similarly, when the features cannot be disambiguated, the likelihood becomes small (Figure~\ref{fig:Reaching}A). This confidence readout also works in other contexts, for instance in the drosophila example frames we only depicted the predicted body parts when the probability was larger than $10\%$ (Figure~\ref{fig:Drosophila}B-D). Using this threshold, the point estimate for the left leg can be automatically excluded in Figure~\ref{fig:Drosophila}C, D. Indeed, all occluded body parts are also omitted in Figure~\ref{fig:Drosophila}B and Figure~\ref{fig:Reaching}F. 

Lastly, once a network is trained on the hand posture frames, the body part locations can be extracted from videos and utilized in many ways. Here we illustrate a few examples: digit positions during a reach across time (Figure \ref{fig:Reaching}B; note that this trajectory comes from frame by frame prediction without any temporal filtering), comparison of movement patterns across body parts (Figure~\ref{fig:Reaching}C, D), dimensionality reduction to reveal the richness of mouse hand postures during reaching (Figure~\ref{fig:Reaching}E) and creating `skeletons' based on semantic meaning of the labels (Figure~\ref{fig:Reaching}F).

\section{Discussion}

Detecting postures from monocular images is a challenging problem. Traditionally, postures are modeled as a graph of parts, where each node encodes the local visual properties of the part in question, and then these parts are connected by spring-like links. This graph is then fit to images by minimizing some appearance cost function~\cite{Felzenszwalb2005}. This minimization is hard to solve, but designing the model topology together with the visual appearance model is even more challenging \cite{Felzenszwalb2005,Toshev2014}; this can be illustrated by considering the diversity of fruit fly (Figure~\ref{fig:Drosophila}) and hand postures we examined (Figure~\ref{fig:Reaching}). In contrast, casting this problem as a minimization with deep residual neural networks allows each joint predictor to have more than just local access to the image~\cite{Toshev2014,Wei16, insafutdinov2016deepercut, pishchulin16cvpr, Feichtenhofer2017}. Due to the extreme depth of ResNets, architectures like DeeperCut have large receptive fields, which can learn to extract postures in a robust way~\cite{insafutdinov2016deepercut}.

Here we demonstrated that cutting-edge deep learning models can be efficiently utilized in the laboratory. Specifically, we leveraged the fact that adapting pre-trained models to new tasks can dramatically reduce the amount of training data required, a phenomena known as transfer learning~\cite{donahue2014decaf,Yosinski2014,Goodfellow2016,insafutdinov2016deepercut,kummerer2016deepgaze}. We first estimated the accuracy of a human labeler who could readily identify the body parts we were interested in for odor guided navigation, and then demonstrated that a deep architecture can achieve similar performance on detection of body parts such as the snout, or the tail after training on only a few hundred images. Moreover, this solution requires no computational body-model, stick figure, time-information, or sophisticated inference algorithm. Thus, it can also be quickly applied to completely different behaviors that pose qualitatively distinct challenges to computer vision, like skilled reaching or egg-laying in drosophila.

We believe that DeepLabCut will supplement the rich literature of computational methods for video analysis~\cite{Drai2001,Sousa2006,Gomez-Marin2012,Matsumoto2013,Machado2015Carey,Guo2015Hantman,Dollar2010,Ben-Shaul2017,Dell2014review,Anderson2014}, where powerful feature detectors of user-defined body parts need to be learned for a specific situation, or where regressors based on standard image features and thresholding heuristics~\cite{Gomez-Marin2012,robie2017machine,Dell2014review} fail to provide satisfying solutions. This is particularly the case in dynamic visual environments (e.g. with varying background and reflective walls (Figures~\ref{fig:TrackingMainFig} and \ref{fig:Drosophila}) or when tracking highly articulated objects like the hand (Figure~\ref{fig:Reaching}). 

\begin{flushleft}
\textbf{Dataset labeling and fine-tuning}
\end{flushleft}

Deep learning algorithms are extremely powerful and can learn to associate arbitrary categories to images~\cite{zhang2016understanding,Goodfellow2016}. This is consistent with our own observation that the training set should be free of errors (Figure~\ref{fig:TrackingMainFig}G) and approximate the diversity of visual appearances. Thus, in order to train DeepLabCut for specific applications we recommend labeling maximally diverse images (i.e. different poses, different individuals, luminance conditions, data collected with different cameras, etc.) in a consistent way and curate the labeled data well. Even for a extremely small training set, the typical errors can be small, but large errors for test images that are quite distinct from the training set can start to dominate the average error. One limitation for generalizing to novel situations comes from stochasticity in training set selection. Given that we only select a small number of training samples (i.e. a few hundred frames), and do so randomly from the data, it is plausible that images representing behaviors that are especially sparse or noisy (i.e. due to motion blur) could be sub-optimally sampled or entirely excluded from the training data, resulting in difficulties at test time. 

Therefore, a user can expand the initial training dataset in an iterative fashion using the score-maps as a guide. Specifically, errors can be addressed via post-hoc ``fine tuning" of the network weights, taking advantage of the fact that the network outputs confidence estimates for its own generated labels (Figure~\ref{fig:Reaching}A, see Methods). By using these confidence estimates to select sequences of frames containing a sparse behavior (by sampling around points of high probability), or to find frames where reliably captured behaviors are largely corrupted with noise (by sampling points of low probability), a user can then selectively label frames based on these confidence criteria to generate a minimal yet additional training set for fine tuning the network. This selectively improves model performance on edge cases, thereby extending the architecture’s capacity for generalization in an efficient way. Thus, for generalization to large data sets, one can use features of the score-maps such as the amplitude or width, or use heuristics such as the continuity of body part trajectories, to identify images where the decoder might make large errors. Images with insufficient automatic labeling performance that are identified in this way can then be manually labeled to increase the training set and iteratively improve the feature detectors. Such an active learning framework can be used to achieve a predefined level of confidence for all images with minimal labeling cost. Then, due to the large capacity of the neural network that underlies the feature detectors, one can continue training the network with these additional examples.

We note however, that not every low probability score-map value necessarily reflects erroneous detection. As we showed, low probabilities can also be indicative of occlusions, as in the case of the digit tips when the mouse is holding the joystick (Figure~\ref{fig:Reaching}). Here, multiple camera angles can be used to fully capture a behavior of interest, or heuristics can be used (such as a body-model) to approximate occluded body parts based on temporal and spatial information. 

\begin{flushleft}
\textbf{Speed and accuracy of DeepLabCut}
\end{flushleft}

Another important feature of DeepLabCut is that it can accurately transform large videos into low-dimensional time sequence data with semantic meaning, as the experimenter pre-selects the parts that will presumably provide the most information about the behavior being studied. Such low-dimensional time sequence data is also highly amenable to behavioral clustering and  analysis due to its computational tractability~\cite{Berman2018,Dell2014review,Anderson2014,gomez2014big}. On modern hardware pose extraction is also fast. For instance, one can process the $682 \times 540$ sized frames of the drosophila behavior at around $30$ frames per second on a NVIDIA $1080$-Ti GPU. Such fast pose extraction can make this tool potentially amenable for real-time feedback~\cite{Mathis2017} based on video-based posture estimates. This processing speed can be further improved by cropping input frames in an adaptive way around the animal, and/or adapting the network architecture to speed up processing times.

\begin{flushleft}
\textbf{Extensions}
\end{flushleft}

As presented, DeepLabCut extracts the posture data frame-by-frame, but one can add temporal Kalman filtering to improve performance (as for other approaches)~\cite{stauffer1999adaptive,Ristic2003beyond,Dell2014review}. Here we omitted such methods due to the high precision of the model without these additional steps, and to focus on the prediction based on single frames solely driven by within-frame visual information  in a variety of contexts. 

While temporal information could indeed be beneficial in certain contexts,  challenges remain to using end-to-end trained deep architectures for video data to extract postures. Due to the curse of dimensionality, deep architectures on videos must rely on lower spatial resolution input images, and thus, the best performing action recognition algorithms still rely on frame-by-frame analysis with deep networks pre-trained on ImageNet due to hardware limitations~\cite{Feichtenhofer2017,Carreira2017,insafutdinov2017cvpr}. As this is an active area of research, we believe this situation will likely change due to improvements in hardware (and in deep learning algorithms), and this should have a strong influence on pose estimation in the future. Therefore currently, in situations where occlusions are very common, such as in social behaviors, pairwise interactions could also be added to improve performance~\cite{Felzenszwalb2005,deChaumont2012,Matsumoto2013, Perez-Escudero2014, Nakamura2016, Uhlmann2017,Dell2014review,insafutdinov2017cvpr,insafutdinov2016deepercut}. Here we focused on the deep feature detectors alone to demonstrate remarkable transfer learning for laboratory tasks without the need for such extensions. 

\begin{flushleft}
\textbf{Conclusions}
\end{flushleft}

Together with this paper we provide an open source software package called DeepLabCut. The toolbox uses the feature detectors from DeeperCut and provides routines to a) extract distinct frames from videos for labeling, b) generate training data based on labels, c) train networks to the desired feature sets, and d) extract these feature locations from unlabeled data (Figure~\ref{fig:toolbox}). The typical use case would be that an experimenter extracts distinct frames from videos and labels the body-parts of interest to create tailored part detectors. Then, after only a few hours of labeling and a day of training the network, they can apply DeepLabCut to novel videos. While we demonstrate the utility of this toolbox on mice and Drosophila, there is no inherent limitation of this framework, and our toolbox can be applied to other model, or non-model, organisms in a diverse range of behaviors.

\section{Methods}

\begin{flushleft}
\textbf{Mouse Odor Trail-Tracking}
\end{flushleft}

 The trail-tracking behavior is part of an investigation into odor guided navigation, where one or multiple wild type (C57BL/6J) mice are running on a paper spool and following odor trails. These experiments were carried out in the Murthy lab at Harvard University and will be published elsewhere. For trail-tracking we extracted $1\,080$ random, distinct frames from multiple experimental sessions observing $7$ different mice. Data was recorded by two different cameras ($640 \times 480$ pixels, and at approximately $1\,700 \times 1\,200$) at $30$ Hz. The latter images are prohibitively large to process by this deep neural network without down-sampling, therefore we cropped around mice to generate images that are approximately $800 \times 800$. One human annotator was instructed to localize the snout, the tip of the left and right ear as well as the base of the tail in the example images on two different occasions, generating two distinct label sets ($>1$ month apart to reduce memory bias, see Figure~\ref{fig:Dataset}).
 
\begin{flushleft}
\textbf{Mouse Reach \& Pull Joystick Task}
\end{flushleft}

Experimental procedures for the training of the joystick behavior and the construction of the behavioral set-up can be found in~\cite{Mathis2017}. In brief, head-fixed mice were trained to reach, grab, and pull a joystick for a liquid reward. To generate a train/test set of images, we labeled $159$ frames at the digit tip, the joint in the middle and at the base of the digit for four digits (which roughly correspond to the proximal interphalangeal joint (PIP) and the metacarpophalangeal joint (MCP), respectively), as well as the base of the hand (wrist). The data was collected across $5$ different mice (C57BL/6J, male and female) and were recorded at $2\,048 \times 1\,088$ resolution with a frame rate of $100 - 320$ Hz. For tracking the digits we used the supplied toolbox code to crop the data to only extract regions of interest containing the movement of the arm to limit the size of the input image to the network.  

All surgical and experimental procedures for mice were in accordance with the National Institutes of Health Guide for the Care and Use of Laboratory Animals and approved by the Harvard Institutional Animal Care and Use Committee. 

\begin{flushleft}
\textbf{Drosophila Egg-Laying Behavior}
\end{flushleft}

Experiments were carried out in the laboratory of Richard Axel at Columbia University and will be published elsewhere. In brief, egg-laying behavior was observed in custom designed 3D-printed chambers (Protolabs). Individual chambers were 4.1 mm deep and tapered from top to bottom, with the top dimensions 7.3 mm $\times$ 5.8 mm and the bottom dimensions 6.7mm $\times$ 4.3mm. One side of the chamber opened to a reservoir within which 1\% agar was poured and allowed to set. Small acrylic windows were slid into place within grooves at the top and bottom to enclose the fly within the chamber and to allow for viewing. The chamber was illuminated by a 2 inch off-axis ring light (Metaphase) and video recording was performed from above the chamber using an IR-sensitive CMOS camera (Basler) with a 0.5x telecentric lens (Edmund Optics) at $20$ Hz ($682 \times 540$ pixels). We identified 12 distinct points of interest to quantify the behavior of interest on the body of the fly. One human annotator manually extracted 589 distinct and informative frames from six different animals, labeling only those features that were visible within a given frame. The 12 points comprise:  four points on the head (the dorsal tip of each compound eye, the ocellus, and the tip of the proboscis), the posterior 10 tip of the scutellum on the thorax, the joint between the femur and tibia on each metathoracic leg, the abdominal stripes on the four most posterior abdominal segments (A3-A6), and the ovipositor.

\begin{flushleft}
\textbf{Deep feature detector architecture}\\
\end{flushleft}

\begin{figure*}
\begin{center}
\includegraphics[width=.9\textwidth]{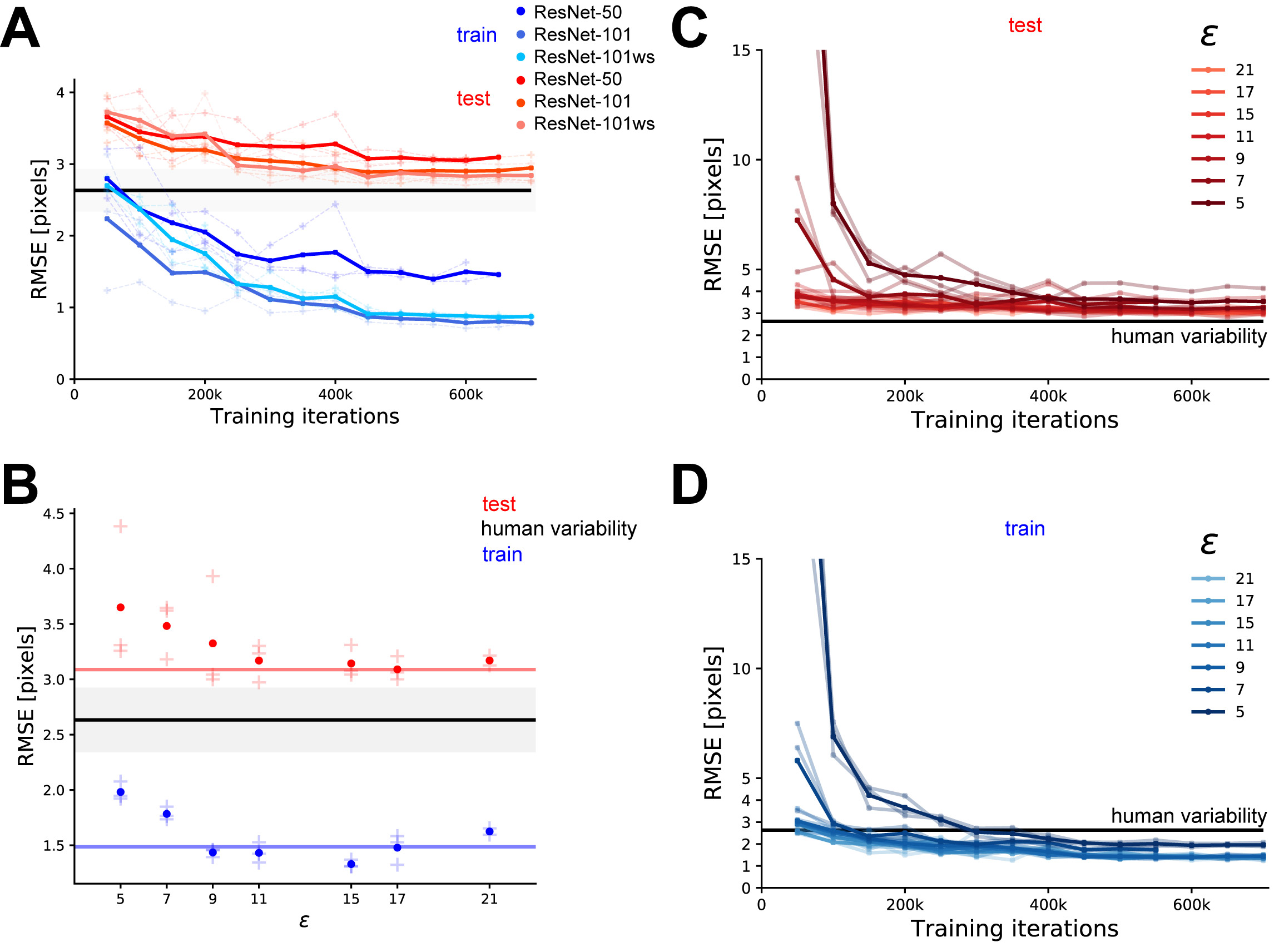}
\end{center}
\caption{{\bf Cross validating model parameters and testing deeper networks}. {\bf A}: Average training and test error for the same $3$ splits with $50\%$ training set size for three different architectures: ResNet-50, ResNet-101 as well as ResNet-101ws, where part loss layers are added to conv4 bank \cite{insafutdinov2017cvpr}. For these networks the training error is strongly reduced, and the test performance modestly improved, indicating that the deeper networks do not overfit (but do not offer radical improvement). Averaged over $3$ splits, individual simulation results shown in as faint lines. The deeper networks reach human level accuracy on test set. The data for ResNet-50 is also depicted in Figure~\ref{fig:TrackingMainFig}D. {\bf B-D}: Cross validating model parameters for ResNet-50 and $50\%$-training set fraction. We varied the distance variable $\epsilon$, which determines the width of the score-map template during training around the ground truth value with scale variable $100\%$ (otherwise the scale ratio of the output layer was set to $80\%$ relative to the the input image size). Varying distance parameters only mildly improves the test performance (after $500k$ training steps). The average performance for scale $0.8$ and $\epsilon=17$ is indicated by horizontal lines (from Figure~\ref{fig:TrackingMainFig}D). In particular, for smaller distance parameters the RMSE increases and learning proceeds much slower (C,D). {\bf C-D}: evolution of the training and test errors at various states of the network training for various distance variables $\epsilon$ corresponding to B.}
\label{fig:Crossvalidation}
\end{figure*}

We employ strong body part detectors, which are part of state-of-the art algorithms for human pose estimation called DeeperCut~\cite{insafutdinov2016deepercut,pishchulin16cvpr,insafutdinov2017cvpr}. Those part detectors build upon state-of-the-art object recognition architectures, namely extremely deep Residual Networks (ResNet)~\cite{He_2016_CVPR}. Specifically, we use a variant of the ResNet with 50 layers, which achieved outstanding performance in object recognition competitions~\cite{He_2016_CVPR}. In the DeeperCut implementation, the ResNets were adapted to represent the images with higher spatial resolution, and the softmax layer used in the original architecture after the conv5 bank (as would be appropriate for object classification) was replaced by “deconvolutional layers” that produce a scalar field of activation values corresponding to regions in original image.  This output is also connected to the conv3 bank to make use of finer features generated earlier in the ResNet architecture ~\cite{insafutdinov2016deepercut}. For each body part, there is a corresponding output layer whose activity represents probability ``score-maps". These score-maps represent the probability that a body part is at a particular pixel~\cite{insafutdinov2016deepercut,pishchulin16cvpr}. During training, a score-map with positive label $1$ (unity probability) is generated for all locations up to $\epsilon$ pixels away from the ground truth per body part (distance variable). The ResNet architecture used to generate features is initialized with weights trained on ImageNet~\cite{He_2016_CVPR} and the cross-entropy loss between the predicted score-map and the ground-truth score-map is minimized by stochastic gradient descent~\cite{insafutdinov2016deepercut}. Around five hundred thousand training steps were enough for convergence and training takes around 24-36h on a GPU (NVIDIA GTX 1080 Ti; see Figure~\ref{fig:TrackingMainFig}B). We used a batch size of $1$, which allows to have images of different size, decreased the learning rate over training and performed data augmentation during training by rescaling the images (all like DeeperCut, but we used a  range of $50\%$ to $150\%$). Unless otherwise noted we used a distance variable $\epsilon=17$ (pixel radius) and scale factor $0.8$ (which affects the ratio of the input image to output score-map). We cross validated the choice of $\epsilon$ for a higher resolution output (scale factor = 1) and found that the test performance is not improved when varying $\epsilon$ widely, but the rate of performance improvement was strongly decreased for small $\epsilon$ (Figure~\ref{fig:Crossvalidation}). We also compared deeper networks with 101 layers, ResNet-101, as well as ResNet-101ws (with intermediate supervision, Figure~\ref{fig:Crossvalidation}A; more technical details for this architecture can be found in~\cite{insafutdinov2016deepercut}). 

\begin{flushleft}
\textbf{Evaluation and error measures}\\
\end{flushleft}

The trained network can be used to predict body part locations. At any state of training the network can be presented with novel frames, for which the prediction of the location of a particular body part is given by the peak of the corresponding score-map. This estimate is further refined based on learned correspondences between the score-map grid and ground truth joint positions~\cite{insafutdinov2017cvpr}. In the case of multiple mice the local maxima of the score-map are extracted as predictions of the body part locations (Figure~\ref{fig:socialmicetracking}).

To compare between datasets generated by the human  scorer, as well as with/between model-generated labellings, we used the Euclidean distance (root mean square error, abbreviated: RMSE) calculated pairwise per body-part. Depending on the context this metric is either shown for a specific body-part, averaged over all body-parts, or averaged over a set of images. To quantify the error across learning, we stored snapshots of the weights in TensorFlow (usually every $50\,000$ iterations) and evaluated the RMSE for predictions generated by these frozen networks post-hoc. Note that the RMSE is not the loss function minimized during training. However, the RMSE is the relevant performance metric for pose estimation for assessing labeling precision.

The RMSE between the first and second annotation is referred to as human variability. In figures we also depict the $95\%$ confidence interval for this RMSE, whose limits are given as mean $\pm 1.96$ times the standard error of the mean (Figure~\ref{fig:TrackingMainFig}C, D, F; Figure~\ref{fig:socialmicetracking} Figure~\ref{fig:bodypartgeneralization}; Figure~\ref{fig:Crossvalidation}A-D). Depending on the figure, the RMSE is averaged over all or just a subset of body parts. 

In (Figure~\ref{fig:Reaching}) we extracted cropped images of the hand from full frames by centering it using the predicted wrist position. We then performed dimensionality reduction by t-SNE embedding of those images~\cite{Pedregosa2011} and randomly selected certain and sufficiently distant points to illustrate the corresponding hand postures. 

\begin{flushleft}
\textbf{DeepLabCut Toolbox}\\
\end{flushleft}

This publication is accompanied by open source Python code for selecting training frames, checking human annotator labels, generating training data in the required format, as well as evaluating the performance on test frames. The toolbox also contains code to extract postures from novel videos with trained feature detectors. Thus, this toolbox allows one to train a tailored network based on labeled images and can then perform automatic labeling for novel data. See \url{https://github.com/AlexEMG/DeepLabCut} for details. 

\vspace{2 mm}

\begin{acknowledgments}
We are grateful to Eldar Insafutdinov and Christoph Lassner for suggestions on how to best use the TensorFlow implementation of DeeperCut. We thank Naoshige Uchida for generously providing resources for the joystick behavior, and Richard Axel for generously providing resources for the drosophila research. We also thank Adrian Hoffmann, Jonas Rauber, Tanmay Nath, David Klindt, and Travis DeWolf for a critical reading of the manuscript as well as members of the Bethge lab, especially Matthias K\"ummerer, for discussions. We also thank the $\beta$-testers for trying our toolbox and sharing their results with us.

\begin{flushleft}
\textbf{Funding:} 
\end{flushleft}
\textbf{AM}: Marie Sklodowska-Curie International Fellowship within the 7th European Community Framework Program under grant agreement No. 622943 and by DFG grant MA 6176/1-1.
\textbf{MWM}: Project ALS (Women $\&$ the Brain Fellowship for Advancement in Neuroscience) \& a Rowland Fellowship from the Rowland Institute at Harvard.
\textbf{MB}: German Science foundation (DFG) through the CRC 1233 on “Robust Vision” and from IARPA through the MICrONS program.

\begin{flushleft}
\textbf{Contributions:} 
\end{flushleft}

Conceptualization: AM, MWM, MB. Software: AM, MWM. Formal Analysis: AM. Experiments: AM \& VNM (trail-tracking), MWM (mouse reaching), KC (Drosophila). Labeling: PM, KC, TA, MWM, AM. Writing: AM \& MWM with input from all authors. 
 
\end{acknowledgments}

%\bibliography{references}
%\begin{thebibliography}{}

%\end{thebibliography}

\newpage
\onecolumngrid
\clearpage

\beginsupplement

\section{Supplementary Materials}
Materials related to main figures of Mathis et al. Please also see \url{http://www.mousemotorlab.org/deeplabcut} for video demonstrations of the tracking.

\end{document}